\documentclass[10pt,twocolumn,letterpaper]{article}

\usepackage{iccv}
\usepackage{times}
\usepackage{epsfig}
\usepackage{graphicx}
\usepackage{amsmath}
\usepackage{amssymb}

\usepackage{color}
\usepackage{booktabs}
\usepackage{multirow}
\usepackage{array}
\usepackage{xcolor}
\usepackage{subcaption}
\usepackage{graphicx}
\usepackage{hhline}
\usepackage{makecell}
\usepackage{colortbl}
\usepackage[normalem]{ulem}
\usepackage{soul}
\usepackage{url}
\usepackage{pifont}
\usepackage{setspace}
\newcommand{\cmark}{\ding{51}}%
\newcommand{\xmark}{\ding{55}}%

\usepackage[square,comma,numbers,sort]{natbib}

\usepackage[pagebackref=true,breaklinks=true,letterpaper=true,colorlinks,bookmarks=false]{hyperref}

\iccvfinalcopy 

\newcommand{\fxf}{\mathbf{x}}
\newcommand{\bPf}{\mathbb{P}}

\DeclareMathOperator*{\argmax}{arg\,max}

\makeatletter
\DeclareRobustCommand\onedot{\futurelet\@let@token\@onedot}
\def\@onedot{\ifx\@let@token.\else.\null\fi\xspace}

\def\eg{\emph{e.g}\onedot} \def\Eg{\emph{E.g}\onedot}
\def\ie{\emph{i.e}\onedot} \def\Ie{\emph{I.e}\onedot}
 
\def\etc{\emph{etc}\onedot} \def\vs{\emph{vs}\onedot}
\def\wrt{w.r.t\onedot}

\makeatother
\newcommand{\thickhline}{%
    \noalign {\ifnum 0=`}\fi \hrule height 1pt
    \futurelet \reserved@a \@xhline
}
\newcolumntype{x}{>\small c}
\newcolumntype{L}[1]{>{\raggedright\let\newline\\\arraybackslash\hspace{0pt}}m{#1}}
\newcolumntype{C}[1]{>{\centering\let\newline\\\arraybackslash\hspace{0pt}}m{#1}}
\newcolumntype{R}[1]{>{\raggedleft\let\newline\\\arraybackslash\hspace{0pt}}m{#1}}

\ificcvfinal\fi
\begin{document}

\title{Spatial Memory for Context Reasoning in Object Detection}

\author{Xinlei Chen \hspace{1.5cm} Abhinav Gupta\\
School of Computer Science, \hspace{0.1cm} Carnegie Mellon University\\
{\tt\small \{xinleic,abhinavg\}@cs.cmu.edu}
}

\maketitle

\begin{abstract}
  Modeling instance-level context and object-object relationships is extremely challenging. It requires reasoning about bounding boxes of different classes, locations \etc. Above all, instance-level spatial reasoning inherently requires modeling conditional distributions on previous detections. Unfortunately, our current object detection systems do not have any {\bf memory} to remember what to condition on! The state-of-the-art object detectors still detect all object in parallel followed by non-maximal suppression (NMS). While memory has been used for tasks such as captioning, they mostly use image-level memory cells without capturing the spatial layout. On the other hand, modeling object-object relationships requires {\bf spatial} reasoning -- not only do we need a memory to store the spatial layout, but also a effective reasoning module to extract spatial patterns. This paper presents a conceptually simple yet powerful solution -- Spatial Memory Network (SMN), to model the instance-level context efficiently and effectively. Our spatial memory essentially assembles object instances back into a pseudo ``image'' representation that is easy to be fed into another ConvNet for object-object context reasoning. This leads to a new sequential reasoning architecture where image and memory are processed in parallel to obtain detections which update the memory again. We show our SMN direction is promising as it provides 2.2\% improvement over baseline Faster RCNN on the COCO dataset so far.
\end{abstract}

\begin{figure}[t]
  \centering
  \includegraphics[width=0.95\linewidth]{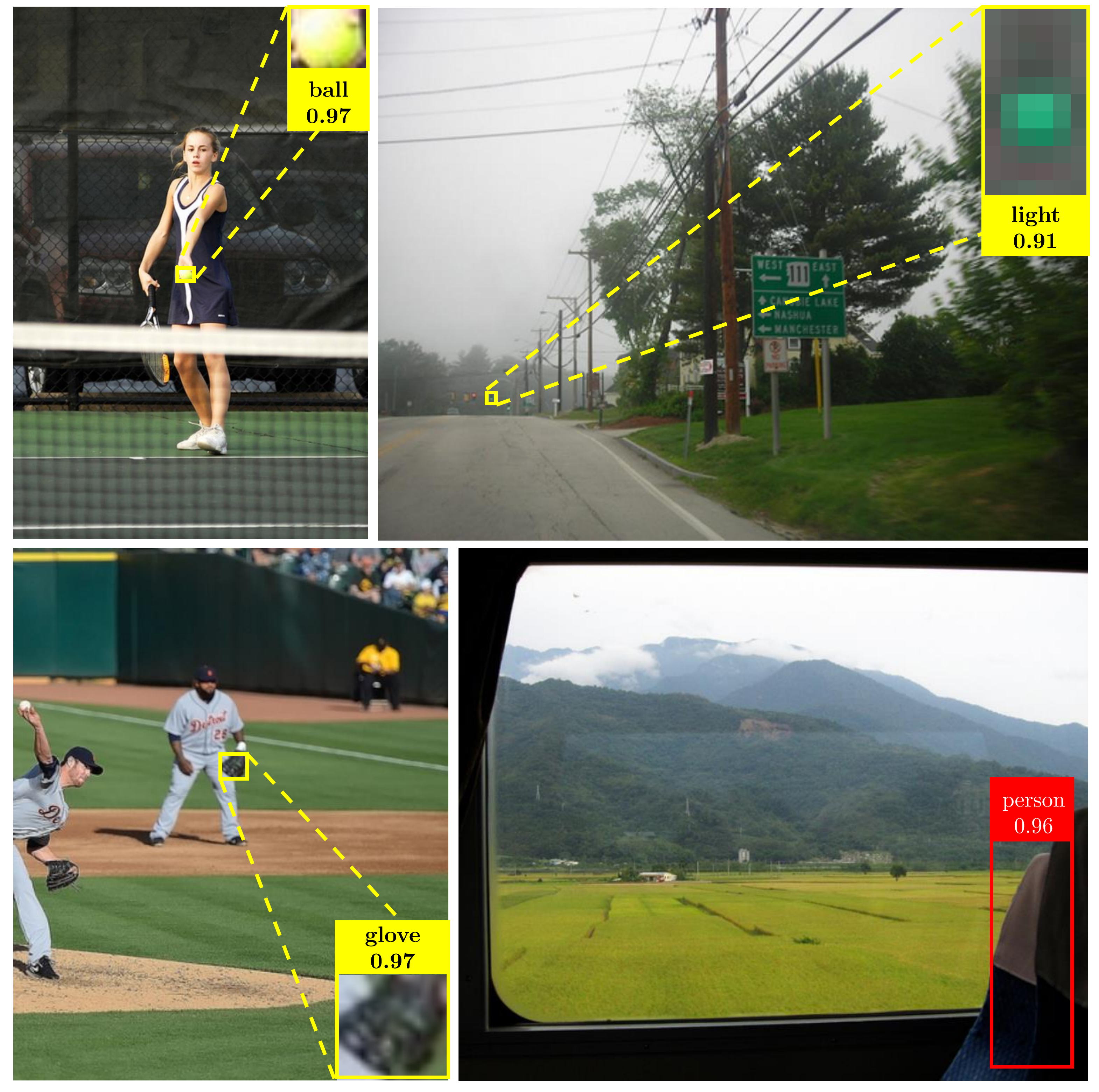}
  \vspace{-0.05in}
  \caption{Evidence of image-level context reasoning inside ConvNets. All examples are from our baseline faster RCNN detector with VGG16 \texttt{conv5\_3} features on COCO~\cite{lin2014microsoft}. Numbers are class confidences. Top and bottom left: three examples where the ConvNet is able to detect tiny and simple-shaped objects much smaller than the receptive field size. Bottom right: a false positive detection for person given the seat on a passenger train.  Our Spatial Memory Network takes advantage of this power by encoding multiple object instances into a ``pseudo'' image representation. \label{fig:teaser}}
  \vspace{-0.2in}
\end{figure}

\vspace{-0.05in}
\section{Introduction\label{sec:intro}}
\vspace{-0.05in}
Context helps image understanding! Apart from strong psychological evidence~\cite{palmer1975effects,hollingworth1998does,bar2004visual,oliva2007role} that context is vital for humans to recognize objects, many empirical studies in the computer vision community~\cite{torralba2003context,carbonetto2004statistical,shotton2006textonboost,tu2010auto,galleguillos2008object,marszalek2009actions,divvala2009empirical,galleguillos2010context,krahenbuhl2011efficient,mottaghi2014role,chen2016deeplab} have also suggested that recognition algorithms can be improved by proper modeling of context. 

But what is the right model for context? Consider the problem of object detection. There are two common models of context often used in the community. The first type of model incorporates image or scene level context~\cite{murphy2003using,torralba2003contextual,hoiem2008putting,malisiewicz2009beyond,li2011extracting,bell2016inside,shrivastava2016contextual}. The second type models object-object relationships at instance-level~\cite{rabinovich2007objects,malisiewicz2009beyond,yao2010modeling,desai2011discriminative,gkioxari2015contextual,gupta2015exploring}. Take the the top-left image of Fig.~\ref{fig:teaser} as an example, both the {\it person} and the {\it tennis racket} can be used to create a contextual prior on where the {\it ball} should be.

Of these two models, which one is more effective for modeling context? A quick glimpse on the current state-of-the-art approaches, the idea of single region classification~\cite{liu2016ssd,ren2015faster,li2016r,google,lin2016feature} with deep ConvNets~\cite{simonyan2014very,he2016deep} is still dominating object detection. On the surface, these approaches hardly use any contextual reasoning; but we believe the large receptive fields of the neurons in fact do incorporate image-level context (See Fig.~\ref{fig:teaser} for evidences). On the other hand, there has been little or no success in modeling object-object relationships or instance-level context in recent years. 

Why so? Arguably, modeling the instance-level context is more challenging. Instance-level reasoning for object detection would have to tackle bounding boxes pairs or groups in different classes, locations, scales, aspect ratios, \etc. Moreover, for modeling image-level context, the grid structure of pixels allows the number of contextual inputs to be reduced efficiently (\eg to a local neighborhood~\cite{shotton2006textonboost,krahenbuhl2011efficient,chen2016deeplab} or a smaller scale~\cite{weinzaepfel2013deepflow,simonyan2014very}), whereas such reductions for arbitrary instances appear to be not so trivial. Above all, instance-level spatial reasoning inherently requires modeling {\it conditional} distributions on previous detections, but our current object detection systems do not have any {\bf memory} to remember what to condition on! Even in the case of multi-class object detection, the joint layout~\cite{desai2011discriminative} is estimated by detecting all objects in parallel followed by non-maximal suppression (NMS)~\cite{felzenszwalb2010object}. What we need is an object detection system with memory built inside it!

Memory has been successfully used in the recognition community recently for tasks such as captioning~\cite{vinyals2014show,donahue2014long,xu2015show,chen2015mind,mao2014deep,venugopalan2015sequence} or visual question answering~\cite{antol2015vqa,zhu2016visual7w,krishna2016visual,gao2015you,Malinowski_2015_ICCV,xiong2016dynamic,andreas2016learning,yang2016stacked,shih2016look,lu2016hierarchical,xu2016ask}. However, these works mostly focus on modeling an image-level memory, without capturing the spatial layout of the understanding so far. On the other hand, modeling object-object relationships requires {\bf spatial} reasoning -- not only do we need a memory to store the spatial layout, but also a suitable reasoning module to extract spatial patterns. This paper presents a conceptually simple yet powerful solution -- Spatial Memory Network (SMN), to model the instance-level context efficiently and effectively. Our key insight is that the best spatial reasoning module is a ConvNet itself! In fact, we argue that ConvNets are actually the most generic\footnote{Many context models can be built or formulated as ConvNets~\cite{zheng2015conditional,xie2016top}.} and effective framework for extracting spatial and contextual information so far! Inspired by this observation, our spatial memory essentially assembles object instances back into a pseudo ``image'' representation that is easy to be fed into another ConvNet to perform object-object context reasoning.

However, if ConvNets are already so excellent at modeling context, why would we even bother something else? Isn't the image itself the ultimate source of information and therefore the best form of ``spatial memory''? Given an image, shouldn't an ultra-deep network already take care of the full reasoning inside its architecture? In spite of these valid concerns, we argue that a spatial memory still presents as an important next step for object detection and other related tasks, for the following reasons:
\begin{itemize}
   \item First, we note that current region-based object detection methods are still treating object detection as a \emph{perception} problem, not a \emph{reasoning} problem: the region classifier still produces multiple detection results around an object instance during inference, and relies on manually designed NMS~\cite{ren2015faster} with a pre-defined threshold for de-duplication. This process can be sub-optimal. We show that with a spatial memory that memorizes the already detected objects, it is possible to learn the functionality of NMS automatically.
   \item Second, replacing NMS is merely a first demonstration for context-based reasoning for object detection. Since the spatial memory is supposed to store both semantic and location information, a legitimate next step would be full context reasoning: \ie, infer the ``what'' and ``where'' of other instances based on the current layout of detected objects in the scene. We show evidence for such benefits on COCO~\cite{lin2014microsoft}.
   \item Third, our spatial memory essentially presents as a general framework to encode instance-level visual knowledge~\cite{li2010object}, which requires the model to properly handle the spatial (\eg overlaps) and semantic (\eg poses) interactions between groups of objects. Our approach follows the spirit of end-to-end learning, optimizing the representation for an end-task -- object detection. Both the representation and the idea can be applied to other tasks that require holistic image understanding~\cite{antol2015vqa,zhu2015building,johnson2016densecap}.
\end{itemize}

\vspace{-0.05in}
\section{Related Work}
\vspace{-0.05in}
As we already mentioned most related work for context and memory in Sec.~\ref{sec:intro}, in this section we mainly review ideas that use sequential prediction for object detection. A large portion of the literature~\cite{lampert2009efficient,gonzalez2015active,lu2016adaptive} focuses on sequential approaches for region proposals (\ie, foreground/background classification). The motivation is to relieve the burden for region classifiers by replacing an exhaustive sliding-window search~\cite{felzenszwalb2010object} with a smarter and faster search process. In the era of ConvNet-based detectors, such methods usually struggle to keep a delicate balance between efficiency and accuracy, since a convolution based $0$/$1$ classifier (\eg region proposal network~\cite{ren2015faster}) already achieves an impressive performance when maintaining a reasonable speed. Sequential search has also been used for localizing small landmarks~\cite{singh2015learning}, but the per-class model assumes the existence of such objects in an image and lacks the ability to use other categories as context. 

Another commonly used trick especially beneficial for reducing localization error is iterative bounding box refinement~\cite{girshick2015fast,ren2015faster,gidaris2016attend,yoo2015attentionnet}, which leverages local image context to predict a better bounding box iteratively. This line of research is complementary to our SMN, since its goal is to locate the original instance \emph{itself} better, whereas our focus is on how to better detect \emph{other} objects given the current detections.

An interesting recent direction focuses on using deep reinforcement learning (DRL) to optimize the sequence selection problem in detection~\cite{caicedo2015active,mathe2016reinforcement,bellver2016hierarchical,liang2017deep}. However, due to the lack of full supervision signal in a problem with high-dimensional action space\footnote{Jointly reason about all bounding boxes and all classes.}, DRL has so far only been used for bounding box refinements or knoledge-assisted detecton, where the action space is greatly reduced. Nevertheless, SMN can naturally serve as an encoder of the state in a DRL system to directly optimize average precision~\cite{henderson2016end}.

Note that the idea of using higher-dimensional memory in vision is not entirely new. It has resemblance to spatial attention, which has been explored in many high-level tasks~\cite{xu2015show,xu2016ask,li2016attentive,ren2016end}. To bypass NMS, LSTM~\cite{hochreiter1997long} cells arranged in 2D order~\cite{stewart2016end} and intersection-over-union (IoU) maps~\cite{hosang2016convnet} have been used for single-class object detection. We also notice a recent trend in using 2D memory as a map for planning and navigation~\cite{gupta2017cognitive,parisotto2017neural}. Our work extends such efforts into generic, multi-class object detection, performing joint reasoning on both space and semantics.

\vspace{-0.05in}
\section{Background: Faster RCNN}
\vspace{-0.05in}
Our spatial memory network is agnostic to the choice of base object detection model. In this paper we build SMN on top of Faster R-CNN~\cite{ren2015faster} (FRCNN) as a demonstration, which is a state-of-the-art detector that predicts and classifies Regions of Interest (RoIs). Here we first give a brief review of the approach.

\subsection{Base Network}
We use VGG16~\cite{simonyan2014very} as the base network for feature extraction. It has $13$ convolutional (\texttt{conv}), $5$ max-pooling (\texttt{pool}), and $2$ fully connected (\texttt{fc}) layers before feeding into the final classifier, and was pre-trained on the ILSVRC challenge~\cite{russakovsky2015imagenet}. Given an image $\mathcal{I}$ of height $h$ and width $w$, feature maps from the last \texttt{conv} layer (\texttt{conv5\_3}) are first extracted by FRCNN. The \texttt{conv5\_3} feature size ($h'$, $w'$) is roughly $\gamma{=}1/16$ of the original image in each spatial dimension. On top of it, FRCNN proceeds by allocating two sub-networks for region proposal and region classification.

\subsection{Region Proposal}
The region proposal network essentially trains a class-agnostic objectness~\cite{alexe2012measuring} classifier, proposing regions that are likely to have a foreground object in a sliding window manner~\cite{felzenszwalb2010object}. It consists of $3$ \texttt{conv} layers, one maps from \texttt{conv5\_3} to a suitable representation for RoI proposals, and two $1{\times}1$ siblings on top of this representation for foreground/background classification and bounding box regression. Note that at each location, anchor boxes~\cite{ren2015faster} of multiple scales ($s$) and aspect ratios ($r$) are used to cover a dense sampling of possible windows. Therefore the total number of proposed
boxes is $K{\approx}h'{\times}w'{\times}s{\times}r$\footnote{Boarder anchors excluded.}. During training and testing, $k{\ll}K$ regions are selected by this network as candidates for the second-stage region classification.

\subsection{Region Classification}
Since the base network is originally an image classifier, region classification network inherits most usable parts of VGG16, with two caveats. First, because RoI proposals can be be arbitrary rectangular bounding boxes, RoI pooling~\cite{girshick2015fast,google} is used in place of \texttt{pool} on \texttt{conv5\_3} to match the the square-sized ($7{\times}7$) input requirement for \texttt{fc6}. Second, the $1,000$-way \texttt{fc} layer for ILSVRC classification is replaced by two \texttt{fc} layers for $C$-way classification and bounding box regression respectively. Each of the $C$ classes gets a separate bounding box regressor. 

\subsection{De-duplication\label{sec:nms}}
We want to point out the often-neglected fact that a standard post-processing step is used in almost all detectors~\cite{felzenszwalb2010object,ren2015faster,liu2016ssd,li2016r} to disambiguate duplications -- NMS. For FRCNN, NMS takes place in both stages. First, for region proposals, it prunes out the overlapping RoIs that are likely corresponding to the same object (``one-for-all-class'') to train the region classifier. Second, for the final detection results, NMS is applied in an isolated, per-class manner (``one-for-each-class''). In this paper, we still use NMS for RoI sampling during training~\cite{chen17implementation}, and mainly focus on building a model to replace the per-class NMS, with the hope that the model can encode the rich interplay across multiple classes when suppressing redundant detections.

\begin{figure}[t]
  \centering
  \includegraphics[width=1.\linewidth]{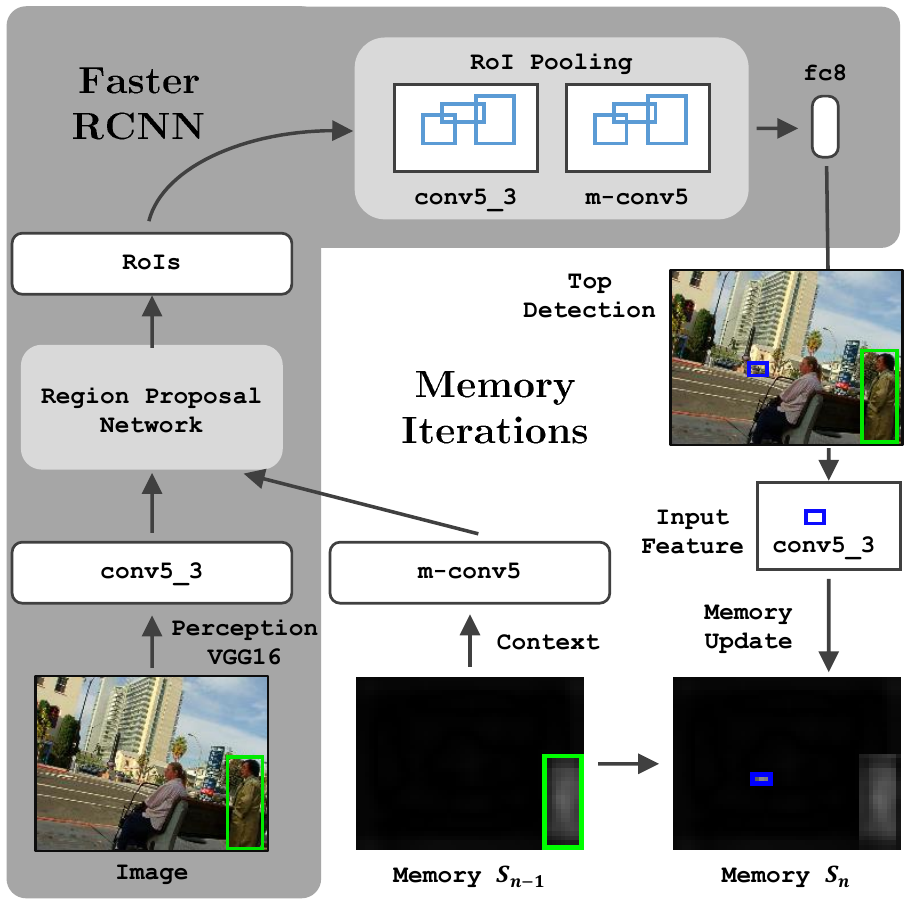}
  \vspace{-0.2in}
  \caption{Overview of memory iterations for object detection. The original components from FRCNN are shown in the gray area. The old detection ({\it person}) is marked with a green box, and the new detection ({\it car}) is marked with blue. Here the network is unrolled one iteration. \label{fig:iter}}
  \vspace{-0.2in}
\end{figure}

\vspace{-0.05in}
\section{Spatial Memory Network}
\vspace{-0.05in}
To better motivate the use of spatial memory network, we resort to a mathematical formulation of the task at hand.
For object detection, the goal is to jointly infer and detect all the object instances $\mathcal{O}{=}[O_1,O_2,O_3,\cdots,O_N]$ given an image $\mathcal{I}$, where $N$ is the maximum number of object instances for any image\footnote{$O_n$ denotes both the class and location of the object instance. When there is not enough foreground objects, the sequence can be padded with the background class.}. Then the objective function of training a model (\eg FRCNN) $\mathcal{M}$ is to maximize the log-likelihood:
\begin{align}
\label{eq:seq}
\argmax_{\mathcal{M}} \mathcal{L} &= \log \bPf(O_{1:N}|\mathcal{M},\mathcal{I})  \nonumber \\
                  &= \sum_{n=1:N}{\log{\bPf\left(O_{n}|\mathcal{O}_{0:n-1},\mathcal{M},\mathcal{I}\right)}},
\end{align}
where $\mathcal{O}_{0:n-1}$ is short for $[O_1,O_2,O_3,\cdots,O_{n-1}]$ and $\mathcal{O}_{0:0}$ is an empty set.
Note that this decomposition of the joint layout probability is exact~\cite{elman1990finding}, regardless of the order we are choosing.

For a region-based object detector, Eq.(\ref{eq:seq}) is approximated by detecting each object instance separately:
\begin{equation}
\label{eq:ind}
\argmax_{\mathcal{M}} \mathcal{L} \approx \sum_{n=1:N}{\log{\bPf\left(O_{n}|\mathcal{M},\mathcal{I}\right)}},
\end{equation}
where NMS shoulders the responsibility to model the correlations in the entire sequence of detections. Since NMS is mostly\footnote{Since NMS is applied in a per-class manner, there is also semantic information.} dependent on overlapping patterns, the information it can provide is limited compared to $\mathcal{O}_{0:n-1}$. 

How can we do better? Inspired by networks that impose a memory~\cite{elman1990finding,hochreiter1997long,chung2014empirical,sukhbaatar2015end,graves2016hybrid} for sequential and reasoning tasks, and the two-dimensional nature of images, we propose to encode $\mathcal{O}_{0:n-1}$ in a spatial memory, where we learn to store all the previous detections. \Ie, we introduce memory variable $\mathcal{S}_{n-1}$, which gets updated each time an object instance is detected, and the approximation becomes: 
\begin{equation}
\label{eq:mem}
\argmax_{\mathcal{M},\mathcal{S}} \mathcal{L} \approx \sum_{n=1:N}{\log{\bPf\left(O_{n}|\mathcal{S}_{n-1}, \mathcal{M},\mathcal{I}\right)}},
\end{equation}
where the memory $\mathcal{S}$ is jointly optimized with $\mathcal{M}$.

With the above formulation, the inference procedure for object becomes conditional: An empty memory is initialized at first (Sec.~\ref{sec:mem}). Once an object instance is detected, selected cells (Sec.~\ref{sec:index}) in the memory gets updated (Sec.~\ref{sec:how}) with features (Sec.~\ref{sec:what}) extracted from the detected region. Then a context model (Sec.~\ref{sec:context}) aggregates spatial and other information from the memory, and outputs (Sec.~\ref{sec:out}) scores that help region proposal and region classification in FRCNN. Then the next potential detection is picked (Sec.~\ref{sec:select}) to update the memory again. This process goes on until a fixed number of iterations have reached (See Fig.~\ref{fig:iter} for an overview).

We now describe each module, beginning with a description of the memory itself.

\subsection{Memory\label{sec:mem}}
Different from previous works that either mixes memory with computation~\cite{elman1990finding,hochreiter1997long,chung2014empirical} or mimics the one-dimensional memory in the Turing machine/von Neumann architecture~\cite{von1993first}, we would like to build a two-dimensional memory for images. This is intuitive because images are intrinsically 2D mappings of the 3D visual world. But more importantly, we aim to leverage the power of ConvNets for context reasoning, which ``forces'' us to provide an image-like 2D input.

How big the memory should be spatially? For object detection, FRCNN that operates entirely on \texttt{conv5\_3} features can already retrieve even tiny objects (\eg the ones in Fig.~\ref{fig:teaser}), suggesting that a resolution $1/16$ of the full image strikes a reasonable balance between speed and accuracy. At each location, the memory cell is a $D{=}256$ dimensional vector that stores the visual information discovered so far. Ideally, the initial values within the memory should capture the photographic bias of a natural image, \ie, prior about where a certain object tend to occur (\eg~{\it sun} is more likely to occur in the upper part). But the prior cannot be dependent on the input image size. To this end, we simply initialize the memory with a fixed spatial size ($20{\times}20{\times}256$ cells), and resize it according to the incoming \texttt{conv5\_3} size using bilinear interpolation. In this way, the memory is fully utilized to learn the prior, regardless of different image sizes.

\subsection{Indexing\label{sec:index}}
The most difficult problem that previous works~\cite{sukhbaatar2015end,graves2016hybrid} face when building an differentiable external memory is the design of memory indexing. The core problem is which memory cell to write to for what inputs. Luckily for our problem, strong correspondence between memory and 2D images solves this problem. Specifically, the target regions to look up in 2D memory are already provided. Furthermore, RoI pooling~\cite{girshick2015fast,google} is precisely the operations needed to \emph{read} off from the spatial memory\footnote{Although RoI pooling only computes partial gradients, back-propagation \wrt bounding box coordinates are not entirely necessary~\cite{ren2015faster} and previously found unstable~\cite{google}.}. The only remaining task is to create a \emph{write} function that updates the memory given a detection. This can be divided into two parts, ``what'' (Sec.~\ref{sec:what}), and ``how'' (Sec.~\ref{sec:how}).

\subsection{Input Features\label{sec:what}}
\begin{figure}[t]
  \centering
  \includegraphics[width=1.\linewidth]{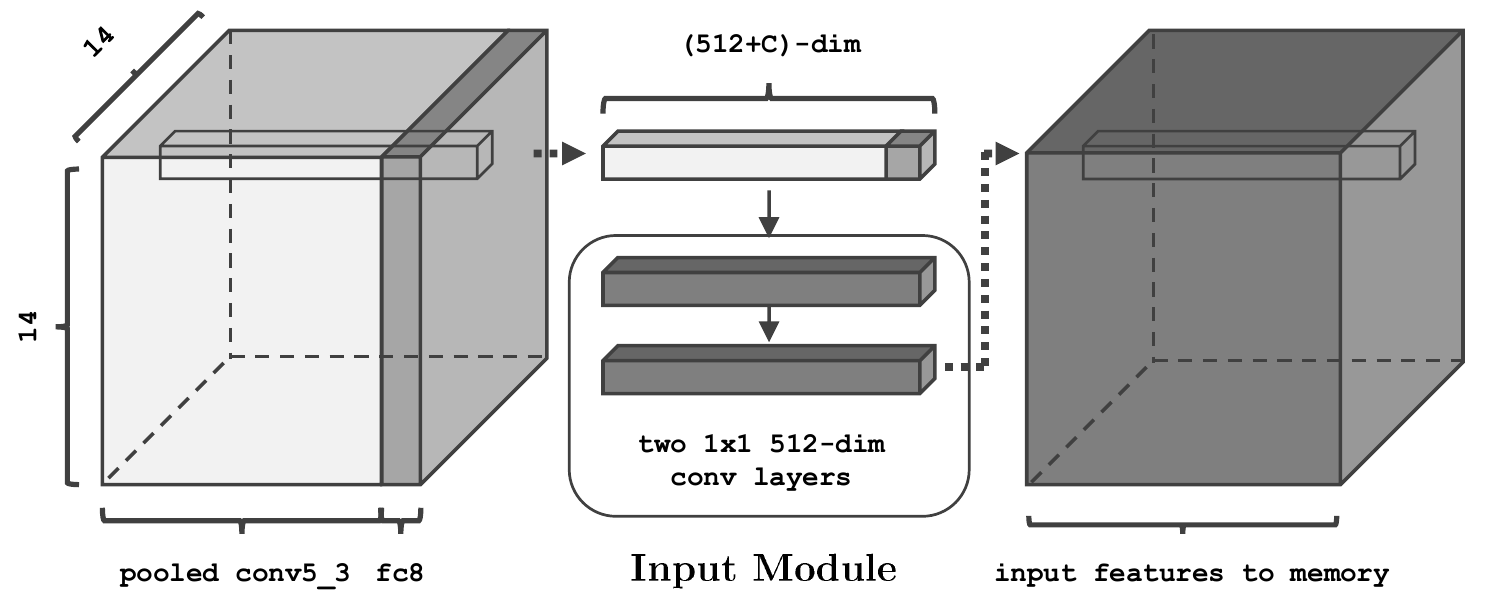}
  \vspace{-0.2in}
  \caption{Illustration of the input module (Sec.~\ref{sec:what}). It assembles spatial and non-spatial features: detection scores after SoftMax (\texttt{fc8}) are tiled at each location of the RoI pooled $14{\times}14$ \texttt{conv5\_3} feature. Two additional \texttt{conv} layers are used to merge the information from two sources. Dotted arrow shows how the feature at one location is transformed. \label{fig:mod}}
  \vspace{-0.2in}
\end{figure}

It may appear trivial, but the decision of what features to insert into the memory requires careful deliberation. First, since \texttt{conv5\_3} feature preserves spatial information, we need to incorporate it. Specifically, we use RoI pooling (without taking Max) to obtain the feature map at the location, and resize it to $14{\times}14$. However, merely having \texttt{conv5\_3} is not sufficient to capture the higher-level semantic information, especially pertaining which object class is detected. The detection score is particularly useful for disambiguation when two objects occur in the same region, \eg, a {\it person} riding a {\it horse}. Therefore, we also include \texttt{fc8} SoftMax score as an input, which is appended at each \texttt{conv5\_3} locations and followed by two $1{\times}1$ \texttt{conv} layers to fuse the information (see Fig.~\ref{fig:mod}). We choose the full score over a one-hot class vector, because it is more robust to false detections. 

\subsection{Writing\label{sec:how}}
Given the region location and the input features $\fxf_n$, we update the corresponding memory cells with a convolutional Gated Recurrent Unit~\cite{chung2014empirical} (GRU), which uses $3{\times}3$ \texttt{conv} filters in place of \texttt{fc} layers as weights. The GRU has a reset gate, and an update gate, shared at each location and activated with Sigmoid function $\sigma(\cdot)$. Hyperbolic tangent $tanh(\cdot)$ is used to constrain the memory values between $-1.$ and $1.$ For alignment, the region from the original memory $\mathcal{S}_{n-1}$ is also cropped with the same RoI pooling operation to $14{\times}14$. After GRU, the new memory cells are placed back to $\mathcal{S}_n$ with a reverse RoI operation.

\subsection{Context Model\label{sec:context}}
Now that the detected objects are encoded in the memory, all we have to do for context reasoning is stacking another ConNet on the top. In the current setup, we use a simple $5$-layer all-convolutional network to extract the spatial patterns. Each \texttt{conv} filter has a spatial size of $3{\times}3$, and channel size of $256$. Padding is added to keep the final layer \texttt{m-conv5} same size of \texttt{conv5\_3}. To ease back-propagation, we add residual connections~\cite{he2016deep} every two layers. 

\subsection{Output\label{sec:out}}
As for the module that outputs the reasoning results, we treat \texttt{m-conv5} exactly the same way as \texttt{conv5\_3} in FRCNN: $3$ \texttt{conv} layers for region proposal, and $2$ \texttt{fc} layers with RoI pooling for region classification. The \texttt{fc} layers have $2048$ neurons each.

\begin{figure}[t]
  \centering
  \includegraphics[width=1.\linewidth]{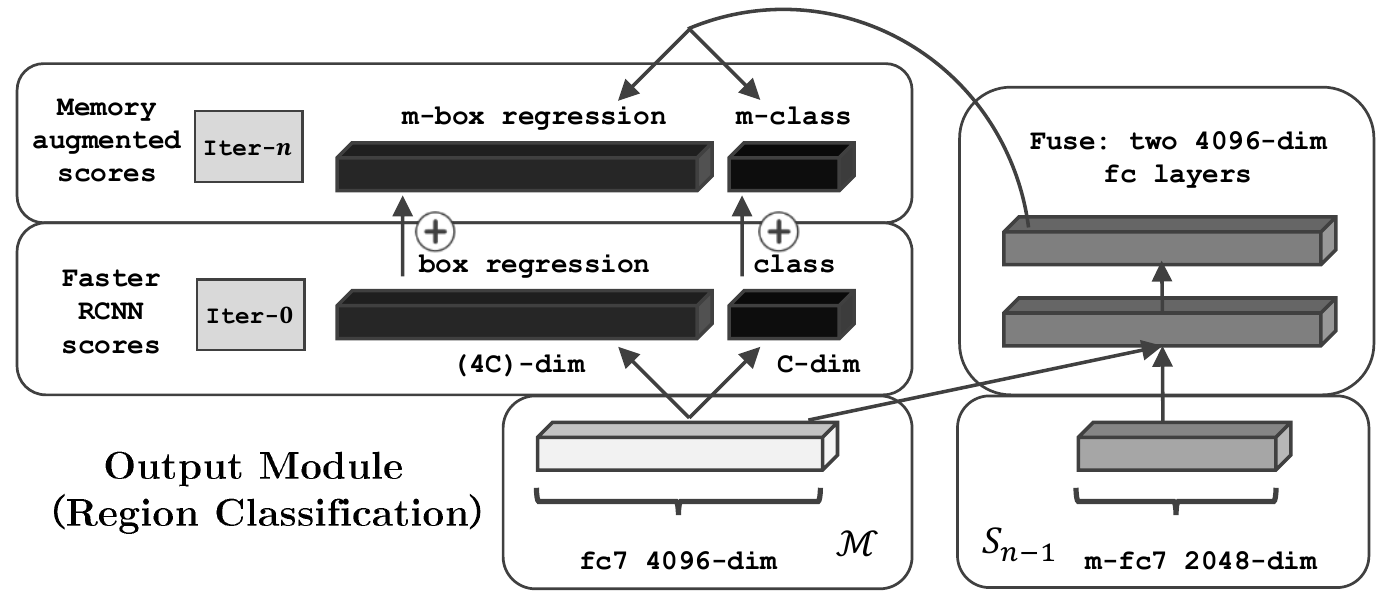}
  \vspace{-0.2in}
  \caption{Illustration of the output module (Sec.~\ref{sec:out}) for region classification. FRCNN scores are optimized at the first iteration when memory is empty, and then augmented with memory scores in later iterations. Same is done for region proposals. Two additional \texttt{fc} layers are used to fuse FRCNN and memory features. \label{fig:mod2}}
  \vspace{-0.2in}
\end{figure}

We design another residual architecture to combine the memory scores with the FRCNN scores (see Fig.~\ref{fig:mod2}): in the first iteration when the memory is empty, we only use FRCNN for detection; from the second iteration on, we add the memory predictions on top of the FRCNN ones, so that the memory essentially provides the additional context to close the gap. This design allows a handy visualization of the prediction difference with/without context. But more importantly, such an architecture is critical to let us converge the full network. Details for this are covered in Sec.~\ref{sec:memaug}.

\subsection{Selecting Next Region\label{sec:select}}
Since spatial memory turns object detection into a sequential prediction problem, an important decision to make is which region to take-in next~\cite{bengio2015scheduled}. Intuitively, some objects are more useful serving as context for others (\eg~{\it person})~\cite{gupta2009observing,yao2010modeling,fouhey2014people,gupta2015exploring}, and some object instances are easier to detect and less prone to consequent errors. However, in this paper we simply follow a greedy strategy -- the most confident foreground object box is selected to update the memory, leaving more advanced models that directly optimize the sequence~\cite{sutton1999policy} as future work. 

\vspace{-0.05in}
\section{Training the Spatial Memory\label{sec:train}}
\vspace{-0.05in}
Like a standard network with recurrent connections, our SMN is trained by back-propagation through time (BPTT)~\cite{williams1995gradient}, which unrolls the network multiple times before executing a weight-update. However, apart from the well-known gradient propagation issue, imposing the conditional structure on object detection incurs new challenges for training. Interestingly, the most difficult one we face in our experiment, is the ``straightforward'' task of de-duplication. 

\begin{figure}[t]
  \centering
  \includegraphics[width=1.\linewidth]{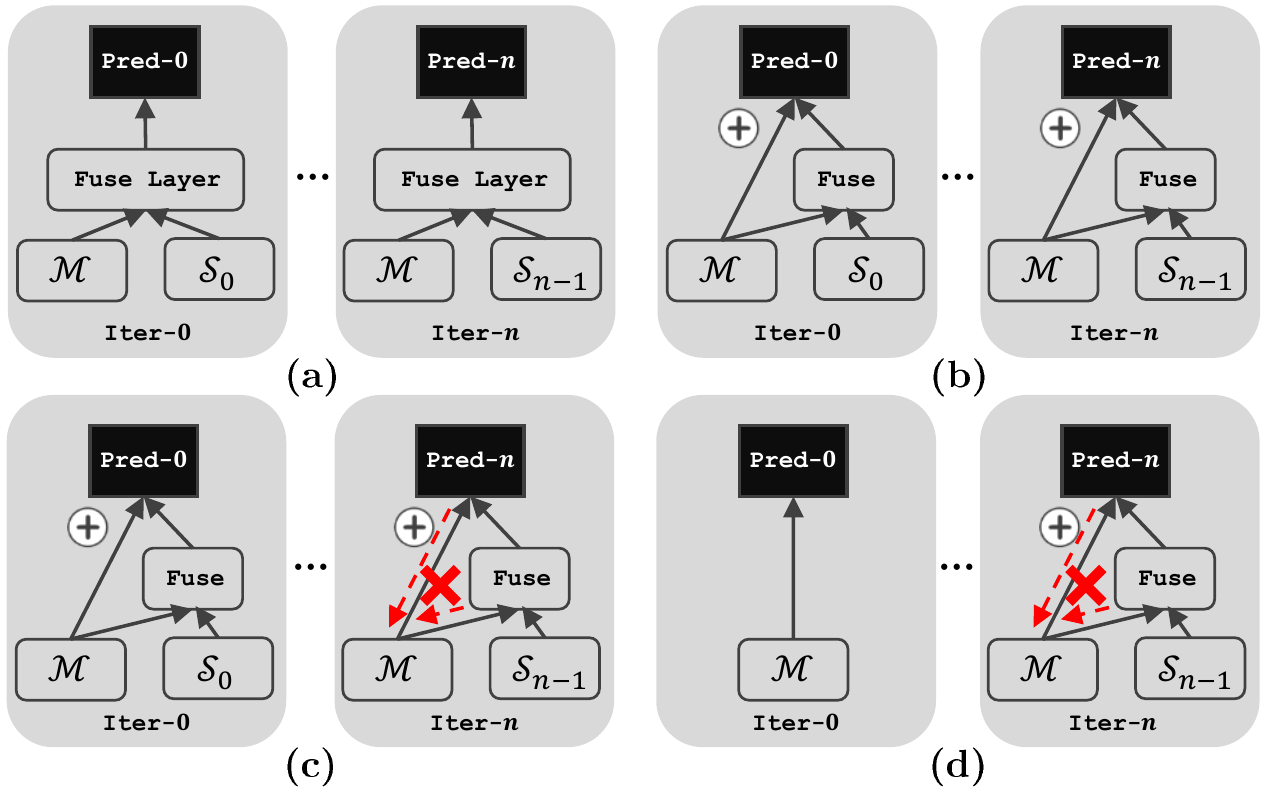}
  \vspace{-0.2in}
  \caption{{Four design choices for learning the functionality of de-duplication. $\mathcal{M}$ is FRCNN features, and $\mathcal{S}_{n-1}$ represents memory features. Each design is shown by two gray panels showing the information flow of Iteration $0$ (left) and Iteration $n{>}0$ (right). We find it hard to even converge the network when the gradient is back-propagated to FRCNN in all iterations (a) \& (b). Stop the gradient in later iterations (c) can successfully converge the network, and our final design (d) separates {\it perception} from {\it reasoning} and makes it easy to visualize the effect of context. All design choices are abstract and apply to both region proposal and classification. Please see Sec.~\ref{sec:memaug} for more details. \label{fig:choice}}}
  \vspace{-0.2in}
\end{figure}

\subsection{Learning De-duplication\label{sec:memaug}}
Simply put, the functionality of de-duplication is: how can the network learn that a detected instance should no longer be detected again? More specifically, we need to design the output module (Sec.~\ref{sec:out}) to fuse the memory ($\mathcal{S}$) and FRCNN ($\mathcal{M}$) beliefs and predict intelligently: when the memory is empty, the FRCNN score should be used; but when the memory has the instance stored, the network needs to ignore, or {\it negate} the cue from FRCNN. 

Since multi-layer networks are universal function approximators~\cite{hornik1989multilayer}, our first attempt is to fuse the information by directly feeding into a multi-layer network (Fig.~\ref{fig:choice} (a)). However, joint-training fails to even converge FRCNN. Suspicious that the longer, weaker supervision might be the cause, we also added skip connections~\cite{bell2016inside} to guide the FRCNN training directly (Fig.~\ref{fig:choice} (b)). Yet it still does not help much. Tracking the learning process, we find where the actual problem lies -- because the network needs to de-duplicate, it keeps receiving contradicting signals: the normal one that guides perception, and the adversarial one that prevents \emph{more} perception. And because $\mathcal{S}$ also starts off from scratch, the signal it can provide is also weak and unreliable. As a result, part of both error signals are back-propagated to $\mathcal{M}$\footnote{Since there are two sets of scores (from $\mathcal{M}$ and fused \texttt{fc}) added together for prediction in Fig.~\ref{fig:choice} (b), we find the conflicting signals are also propagated to the \emph{biases} of these predictions: resulting in one going up and the other down while essentially canceling each other.}, causing trouble for learning further.

Realizing where the issue is, a direct solution is to just stop the adversarial signal from flowing back and canceling the normal one. Therefore, we stopped the gradient to FRCNN from second iteration on (Fig.~\ref{fig:choice} (c)), and the network can successfully converge.

To make it easy for training and showing the confidence changes for consequent detections given the context, we further reduced the architecture to exclude all memory related weights in the first iteration (Fig.~\ref{fig:choice} (d)). This way, the change in predictions with/without memory can be read-off directly\footnote{Otherwise we have to run the inference again with $\mathcal{S}_{0}$.}, and training can be done separately for $\mathcal{M}$ and $\mathcal{S}$.

\subsection{RoI Sampling\label{sec:flip}}
To avoid getting overwhelmed by negative boxes, FRCNN enforces a target sampling ratio for foreground/background boxes. The introduction of a spatial memory that learns to de-duplicate, brings in another special type -- regions whose label is flipped from previous iterations. To keep these regions from being buried in negative examples too, we changed the sampling distribution to include flipped regions.

It is important to point out that RoI sampling greatly enhances the robustness of our sequential detection system. Because only $k{\ll}K$ regions are sampled from all regions, the overall most confident RoI is not guaranteed to be picked when updating the memory. This opens up chances for other highly confident boxes to be inserted into the sequence as well~\cite{sutton1999policy} and reduces over-fitting.

\vspace{-0.05in}
\subsection{Multi-Tasking\label{sec:mult}}
\vspace{-0.05in}

We also practiced the idea of multi-task learning for SMN. The major motivation is to force the memory to memorize more: the basic SMN is only asked fulfill the mission of predicting the missing objects, which does not necessarily translate to a good memorization of previously detected objects. \Eg, it may remember that one region has an object in general, but does not store more categorical information beyond that. To better converge the memory, we also added a \emph{reconstruction} loss~\cite{chen2015mind,rohrbach2016grounding}, \ie, letting the network in addition predict the object classes it has stored in the memory. Specifically, we add an identical set of branches on top of the \texttt{m-conv5} features as FRCNN, for both region proposal and region classification in each iteration. These weights are used to predict only the previously detected objects. 

\vspace{-0.05in}
\subsection{Stage-wise Training\label{sec:stage}}
\vspace{-0.05in}
Thanks to the design of our memory augmented prediction, so far we have trained the full model in two separate stages, where FRCNN $\mathcal{M}$, the \emph{perception} model can be optimized independently at first; then the \emph{reasoning} model with spatial memory $\mathcal{S}$ is learned on top of fixed $\mathcal{M}$. This helps us isolate the influence of the base model and focus directly on the study of SMN. 

For efficiency, we also follow a curriculum learning~\cite{bengio2009curriculum} strategy: bootstrap a SMN of more iterations (\eg $N{=}10$) with a pre-trained SMN of fewer iterations (\eg $N{=}5$). As $N$ gets larger, the task becomes harder. Curriculum learning does not require re-learning de-duplication (which we learn with $N$ from $2$ to $4$), and allows the network to focus more on object-object relationships instead.

\subsection{Hyper-parameters}
Given a pre-trained FRCNN or SMN (in the case of curriculum learning), we train a fixed number of $30$k steps. The initial learning rate is set to $1e{-3}$ and reduced to $1e{-4}$ after $20$k steps. Since we do not use automatic normalization tricks~\cite{ioffe2015batch,liu2015parsenet}, different variances are manually set when initializing weights from scratch, in order to let different inputs contribute comparably (\eg when concatenating \texttt{fc7} and \texttt{m-fc7}).
Other hyper-parameters are kept the same to the ones used in FRCNN.

\vspace{-0.05in}
\section{Experimental Results}
\vspace{-0.05in}
We highlight the performance of our spatial memory network on COCO~\cite{lin2014microsoft}. However, for ablative analysis and understanding the behaviour of our system, we use both PASCAL VOC 2007~\cite{everingham2010pascal} and COCO~\cite{lin2014microsoft}. For VOC we use the \textit{trainval} split for training, and \textit{test} for evaluation. For COCO we use \textit{trainval35k}~\cite{bell2016inside} and \textit{minival}. For evaluation, toolkits provided by the respective dataset are used. The main metrics (mAP, AP and AR) are based on detection average precision/recall.

\noindent {\bf Implementation Details:} We use TensorFlow to implement our model, which is built on top of the open-sourced FRCNN implementation\footnote{\url{https://github.com/endernewton/tf-faster-rcnn}} serving as a baseline. For COCO, this implementation has an AP of $29.1\%$ compared to the original one $24.2\%$~\cite{ren2015faster}.

Original FRCNN uses NMS for region sampling as well. However, NMS hurts our performance more since we do sequential prediction and one miss along the chain can negatively impact all the follow-up detections. To overcome this disadvantage, we would ideally like to examine all $K$ regions in a sliding window fashion. However, due to the GPU memory limit, the top $5$k regions are used instead. We analyze this choice in ablative analysis (Sec.~\ref{sec:abla}). Due to the same limitation, our current implementation of SMN can only unroll $N{=}10$ times in a single GPU. At each timestep in SMN, we do a soft max-prediction for the top box selected, so that a single box can be assigned to multiple classes. We will also justify and analyze this choice in Sec.~\ref{sec:abla}.

\noindent {\bf Initial Results:} Table~\ref{tab:initial} shows the initial results of our approach as described. As it can be seen for $N{=}5$ detections per image our SMN give an AP of $24.5\%$ and for $N{=}10$ if gives an AP of $28.1\%$. When the baseline is allowed the same number of detections ($N{=}5,10$), the AP is $23.8\%$ and $27.1\%$. Therefore, while we do outperform baseline for fixed number of detections per image, due to limited roll-out capability we are still ${\sim}1\%$ below the baseline~\cite{chen17implementation}.

\begin{table}[t]
\centering
\renewcommand{\arraystretch}{1.1}
\renewcommand{\tabcolsep}{1.2mm}
\caption[caption]{\label{tab:initial}{Baseline and initial analysis on COCO 2014 minival when constraining the number of detections $N{=}5$/$10$. AP and AR numbers are from COCO evaluation tool.}}
\vspace{-0.1in}
\resizebox{0.99\linewidth}{!}{
\begin{tabular}{@{} C{0.5cm} !{\color{gray}\vrule} L{3.5cm} !{\color{gray}\vrule} x !{\color{gray}\vrule} x !{\color{gray}\vrule} x x x x @{}}
\Xhline{1pt}
$N$ & \textbf{Method} & AP & AR-10 & AR-S & AR-M & AR-L \\
\Xhline{1pt}
- & FRCNN~\cite{ren2015faster} & 24.2 & 33.7 & 11.7 & 39.5 & 54.1 \\
- & Baseline~\cite{chen17implementation} & 29.1 & 38.7 & 17.7 & 44.9 & 56.9 \\
\Xhline{0.5pt}
\parbox[t]{2.5mm}{\multirow{2}{*}{\rotatebox[origin=c]{90}{\footnotesize $N{=}5$}}} & Baseline & 23.8 & 27.8 & 7.0 & 28.7 & 48.4 \\
& SMN & {\bf 24.5} & {\bf 28.9} & {\bf 7.3} & {\bf 29.7} & {\bf 50.6} \\

\Xhline{0.5pt}
\parbox[t]{2.5mm}{\multirow{2}{*}{\rotatebox[origin=c]{90}{\footnotesize $N{=}10$}}} & Baseline & 27.1 & 33.5 & 10.8 & 36.7 & 53.8 \\
& SMN & {\bf 28.1} & {\bf 35.0} & {\bf 11.5} & {\bf 38.1} & {\bf 56.4} \\
\Xhline{1pt}
\end{tabular}
 }
\vspace{-0.2in}
\end{table}

\subsection{SMN for Hard Examples\label{sec:moreiter}}
In this section, we want to go beyond $N{=}10$ detections and see if the overall detection performance can be improved with SMN. Intuitively, for highly confident detections, ConvNet-based FRCNN is already doing a decent job and not much can be learned from an additional memory. It is the ``tails'' that need help from the context! This means two things: 1) with a limited resource budget, SMN should be used in later iterations to provide conditional information; and 2) at the beginning of the sequence, a standard FRCNN can work as a proxy. Given these insights, we experimented with the following strategy: For the first $N_1$ iterations, we use a standard FRCNN to detect easier objects and feed the memory with a sequence ordered by FRCNN confidence (after per-class NMS). Memory gets updated as objects come in, but does not output features to augment prediction. Only for the later $N_2$ iterations it acts normally as a context provider to detect harder examples. For COCO, we set $N_1{=}50$ and bootstrap from a $N_2{=}10$ SMN model.

Although SMN is trained with the goal of context reasoning and learns new functionality (\eg de-duplication) that the original FRCNN does not have, it does have introduced more parameters for memory-augmented prediction. Therefore, we also add a MLP baseline, where a $5$-layer ConvNet (Sec.~\ref{sec:context}) is directly stacked on top of \texttt{conv5\_3} for context aggregation, and the same output modules (Sec.~\ref{sec:out}) are used to make predictions.

The results can be found in Table~\ref{tab:final}. As can be seen, on our final system, we are $2.2\%$ better than the baseline FRCNN. This demonstrates our ability to find hard examples. It is worth noting that here {\it hard} does not necessarily translate to {\it small}. In fact, our reasoning system also helps big objects, potentially due to its ability to perform de-duplication more intelligently and benefit larger objects that are more likely to overlap.

\begin{table}[t]
\centering
\renewcommand{\arraystretch}{1.1}
\renewcommand{\tabcolsep}{1.2mm}
\caption[caption]{\label{tab:final}{Final comparison between SMN and baselines. We additionally include MLP baseline where the number of parameters are kept the same as SMN for context aggregation and output. Top $5$k regions are used to select proposal instead of NMS.}}
\vspace{-0.1in}
\resizebox{0.99\linewidth}{!}{
\begin{tabular}{@{} L{2.0cm} !{\color{gray}\vrule} x !{\color{gray}\vrule} x !{\color{gray}\vrule} *{8}{x} @{}}
\Xhline{1pt}
\textbf{Method} & AP & AP-.5 & AP-.75 & AP-S & AP-M & AP-L & AR-S & AR-M & AR-L \\
\Xhline{1pt}
Baseline~\cite{chen17implementation} & 29.4 & 50.0 & 30.9 & 12.2 & 33.7 & 43.8 & 18.5 & 45.5 & 58.9 \\
\Xhline{0.5pt}
MLP & 30.1 & 50.8 & 31.7 & 12.5 & 34.2 & 44.5 & 19.2 & 47.0 & 59.8 \\
SMN & {\bf 31.6} & {\bf 52.2} & {\bf 33.2} & {\bf 14.4} & {\bf 35.7} & {\bf 45.8} & {\bf 20.5} & {\bf 48.8} & {\bf 63.2} \\
\Xhline{1pt}
\end{tabular}
 }
\vspace{-0.1in}
\end{table}

\begin{figure}[t]
  \centering
  \includegraphics[width=1.\linewidth]{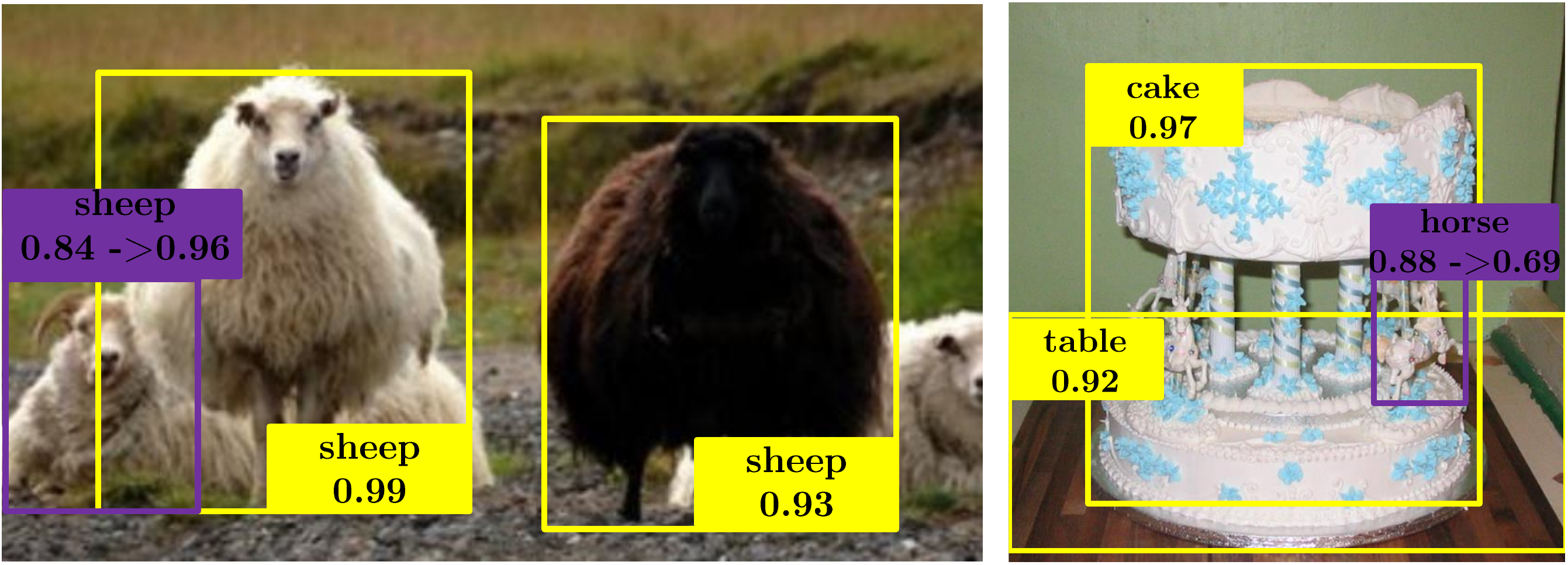}
  \vspace{-0.2in}
  \caption{{Examples of context has helped improve scores by reasoning. Left: the score of {\it sheep} is increased due to presence of other {\it sheep} in background. Right: the score of {\it horse} is decreased due to the detection of {\it cake} and {\it table}. \label{fig:prior}}}
  \vspace{-0.1in}
\end{figure}

\noindent {\bf Qualitative Results:} We show a couple of examples of how context using spatial memory can help improve the performance and detections. In the first case, the score of {\it sheep} gets boosted due to other {\it sheep}. The score of {\it horse} decreases due to the detection of {\it cake} and {\it table}. Please check the supplementary material for more examples.

\vspace{-0.05in}
\subsection{Ablative Analysis~\label{sec:abla}}
\vspace{-0.05in}

We now perform ablative analysis to explain all our choices for the final implementation. For ablative analysis, we use both VOC and COCO datasets. The numbers are summarized in Table~\ref{tab:ablative}. For the comparisons shown here, we switch back to the standard NMS-based region sampling and select top $k{=}300$ RoIs as in original FRCNN. Also, when we do the roll-out, at each step we choose one detection and perform HardMax (rather than SoftMax): make the hard decision about what class does the selected box belong to -- a natural idea for sequential prediction.

For $N{=}5$, we compared three models. First, SMN Base, where we simply train the network as is done in FRCNN. Next, regions with flipped labels (Sec.~\ref{sec:flip}) are added to replaces some of the negative example -- for training region proposal the ratio for positive/flipped/negative is $2{:}1{:}1$, and for region classification it is $1{:}1{:}2$. Third, SMN Full, where we keep the previous sampling strategy and in addition include the reconstruction loss (Sec.~\ref{sec:mult}). Overall, both strategies help performance but with a seemly different strength: sampling flipped regions helps more on small objects, and multi-task learning helps more on bigger ones.

However, our best performance in Table~\ref{tab:ablative} is still behind the baseline and judging from the COCO AR we believe the biggest issue lies in recall. Therefore, we take the best SMN Full model and conduct two other investigations specifically targeting recall. Here we only list the final results, please see supplementary material for more discussions. 

\begin{table}[t]
\centering
\renewcommand{\arraystretch}{1.1}
\renewcommand{\tabcolsep}{1.2mm}
\caption[caption]{\label{tab:ablative}{Ablative analysis on VOC 2007 test and COCO 2014 minival. All approaches constrained by detections $N{=}5$/$10$. mAP is used to evaluate VOC, AP and AR numbers are from COCO.}}
\vspace{-0.1in}
\resizebox{0.99\linewidth}{!}{
\begin{tabular}{@{} C{0.5cm} !{\color{gray}\vrule} L{3.5cm} !{\color{gray}\vrule} x !{\color{gray}\vrule} x !{\color{gray}\vrule} x x x x @{}}
\Xhline{1pt}
$N$ & \textbf{Method} & mAP & AP & AR-10 & AR-S & AR-M & AR-L \\
\Xhline{1pt}
\parbox[t]{2.5mm}{\multirow{4}{*}{\rotatebox[origin=c]{90}{\footnotesize $N{=}5$}}} & Baseline (FR-CNN) & {\bf 65.8} & 23.6 & 27.6 & {\bf 7.0} & {\bf 29.1} & {\bf 47.4} \\
& SMN Base & 63.6 & 23.3 & 27.2 & 6.7 & 28.0 & 46.1 \\
& ~~~~~~$+$~Sample Flipped & 64.4 & 23.5 & 27.2 & 6.9 & 28.4 & 46.4 \\
& SMN Full & 64.6 & {\bf 23.8} & {\bf 27.7} & 6.9 & 28.5 & {\bf 47.4} \\
\Xhline{0.5pt}
\parbox[t]{2.5mm}{\multirow{3}{*}{\rotatebox[origin=c]{90}{\footnotesize $N{=}10$}}} & Baseline & {\bf 70.3} & 26.9 & {\bf 33.2} & {\bf 10.9} & {\bf 36.6} & {\bf 52.7} \\
& SMN Full & 67.5 & 26.6 & 32.6 & 10.3 & 35.6 & 52.1 \\
& ~~~~~~$+$Tune from $N{=}5$ & 67.8 & {\bf 27.1} & 32.7 & 10.3 & 35.9 & 52.3 \\
\Xhline{1pt}
\end{tabular}
 }
\vspace{-0.1in}
\end{table}

\begin{table}[t]
\centering
\renewcommand{\arraystretch}{1.1}
\renewcommand{\tabcolsep}{1.2mm}
\caption[caption]{\label{tab:recall}{Investigating the recall issue. {\bf S} stands for SoftMax based testing, and {\bf H} for HardMax. {\bf $\not$N} is short for Non-aggressive NMS, where top $5$k RoIs are directly selected without NMS.}}
\vspace{-0.1in}
\resizebox{0.99\linewidth}{!}{
\begin{tabular}{@{} C{0.5cm} !{\color{gray}\vrule} L{1.5cm} !{\color{gray}\vrule} C{0.8cm} !{\color{gray}\vrule} C{0.8cm} !{\color{gray}\vrule} x !{\color{gray}\vrule} x !{\color{gray}\vrule} x x x x @{}}
\Xhline{1pt}
$N$ & \textbf{Method} & \textbf{$\not$N} & \textbf{Max} & mAP & AP & AR-10 & AR-S & AR-M & AR-L \\
\Xhline{1pt}
\parbox[t]{2.5mm}{\multirow{6}{*}{\rotatebox[origin=c]{90}{\footnotesize $N{=}5$}}} & Baseline & \xmark & {\bf S} & 65.8 & 23.6 & 27.6 & 7.0 & 29.1 & 47.4 \\
& SMN Full & \xmark & {\bf S} & 66.4 & 24.1 & 28.8 & {\bf 7.5} & {\bf 29.7} & 50.0 \\
\Xcline{2-10}{0.5pt}
& Baseline & \xmark & {\bf H} & 65.4 & 23.5 & 27.2 & 6.7 & 28.6 & 46.9 \\
& SMN Full & \xmark & {\bf H} & 64.6 & 23.8 & 27.7 & 6.9 & 28.5 & 47.4 \\
\Xcline{2-10}{0.5pt}
& Baseline & \cmark & {\bf S} & 66.0 & 23.8 & 27.8 & 7.0 & 28.7 & 48.4 \\
& SMN Full & \cmark & {\bf S} & {\bf 66.6} & {\bf 24.5} & {\bf 28.9} & 7.3 & {\bf 29.7} & {\bf 50.6} \\
\Xhline{0.5pt}
\parbox[t]{2.5mm}{\multirow{6}{*}{\rotatebox[origin=c]{90}{\footnotesize $N{=}10$}}} & Baseline & \xmark & {\bf S} & 70.3 & 26.9 & 33.2 & 10.9 & 36.6 & 52.7 \\
& SMN Full & \xmark & {\bf S} & 69.4 & 27.7 & {\bf 35.0} & {\bf 11.6} & 37.6 & 55.7 \\
\Xcline{2-10}{0.5pt}
& Baseline & \xmark & {\bf H} & 68.0 & 26.4 & 31.9 & 9.7 & 35.0 & 50.7 \\
& SMN Full & \xmark & {\bf H} & 67.8 & 27.1 & 32.7 & 10.3 & 35.9 & 52.3 \\
\Xcline{2-10}{0.5pt}
& Baseline & \cmark & {\bf S} & {\bf 70.4} & 27.1 & 33.5 & 10.8 & 36.7 & 53.8 \\
& SMN Full & \cmark & {\bf S} & 70.0 & {\bf 28.1} & {\bf 35.0} & 11.5 & {\bf 38.1} & {\bf 56.4} \\
\Xhline{1pt}
\end{tabular}
 }
\vspace{-0.2in}
\end{table}

\noindent {\bf SoftMax \vs HardMax:} First, we address a subtle question: if we take top $N$ detections with the memory and compare them directly with top $N$ detections of Faster R-CNN: are these results comparable?  It turns out to be not! As mentioned in Sec.~\ref{sec:nms}, because NMS is applied in a per-class manner, the actual number of box candidates it can put in the final detection is $k{\times}C$. To make it more clear, for a confusing region where \eg the belief for {\it laptop} is $40\%$ and {\it keyboard} is $35\%$, NMS can keep both candidates in the top $N$ detections, whereas for SMN it can only keep the maximum one\footnote{It's also a result of our current input feature design, where we only used \texttt{fc8} and \texttt{conv5\_3} features to update the memory without a top-down notion~\cite{tdm_cvpr17} of which class is picked, so there's no more need for SMN to return.}. Therefore, to be fair, we try: a) HardMax for baseline; and b) SoftMax for SMN.

\noindent {\bf Non-aggressive NMS:} Finally, we also evaluate our choice of non-aggressive NMS during RoI sampling. Both baseline and SMN perform better with $5$k proposals; however our boost on AP is more significant due to sequential prediction issues.

\vspace{-0.05in}
\section{Conclusion}
\vspace{-0.05in}
This paper is our first step towards instance-level reasoning in object detection with ConvNets. We introduce a simple yet powerful framework of spatial memory network, to model the instance-level context efficiently and effectively. Our spatial memory essentially assembles object instances back into a pseudo ``image'' representation. This memory can simply be fed into another ConvNet to extract context information and perform object-object relationship reasoning. We show our SMN direction is promising as it provides $2.2\%$ improvement over baseline Faster RCNN on the COCO dataset so far. We believe our framework is generic and should promote research focusing on knowledge-based reasoning on images.

\vspace{-0.05in}
\section*{Acknowledgements}
\vspace{-0.05in}
XC would like to thank Jiwei Li for helpful discussions on reinforcement learning, and everyone who took time to provide useful feedback. This research is supported in part by ONR MURI N000141612007, the Office of the Director of National Intelligence (ODNI), Intelligence Advanced Research Projects Activity (IARPA) and hardware
donations from NVIDIA. The views and conclusions contained hereon are those of the authors and should not be interpreted as necessarily representing the official policies, either expressed or implied of ODNI, IARPA or the US government.
The US Government is authorized to reproduce and distribute the
reprints for governmental purposed notwithstanding any copyright annotation
thereon. 

{\small
\bibliographystyle{ieee}
\bibliography{smm}
}

\newpage

\section*{A1. More Qualitative Results\label{sec:more}}
\begin{figure}[t]
  \centering
  \includegraphics[width=1.\linewidth]{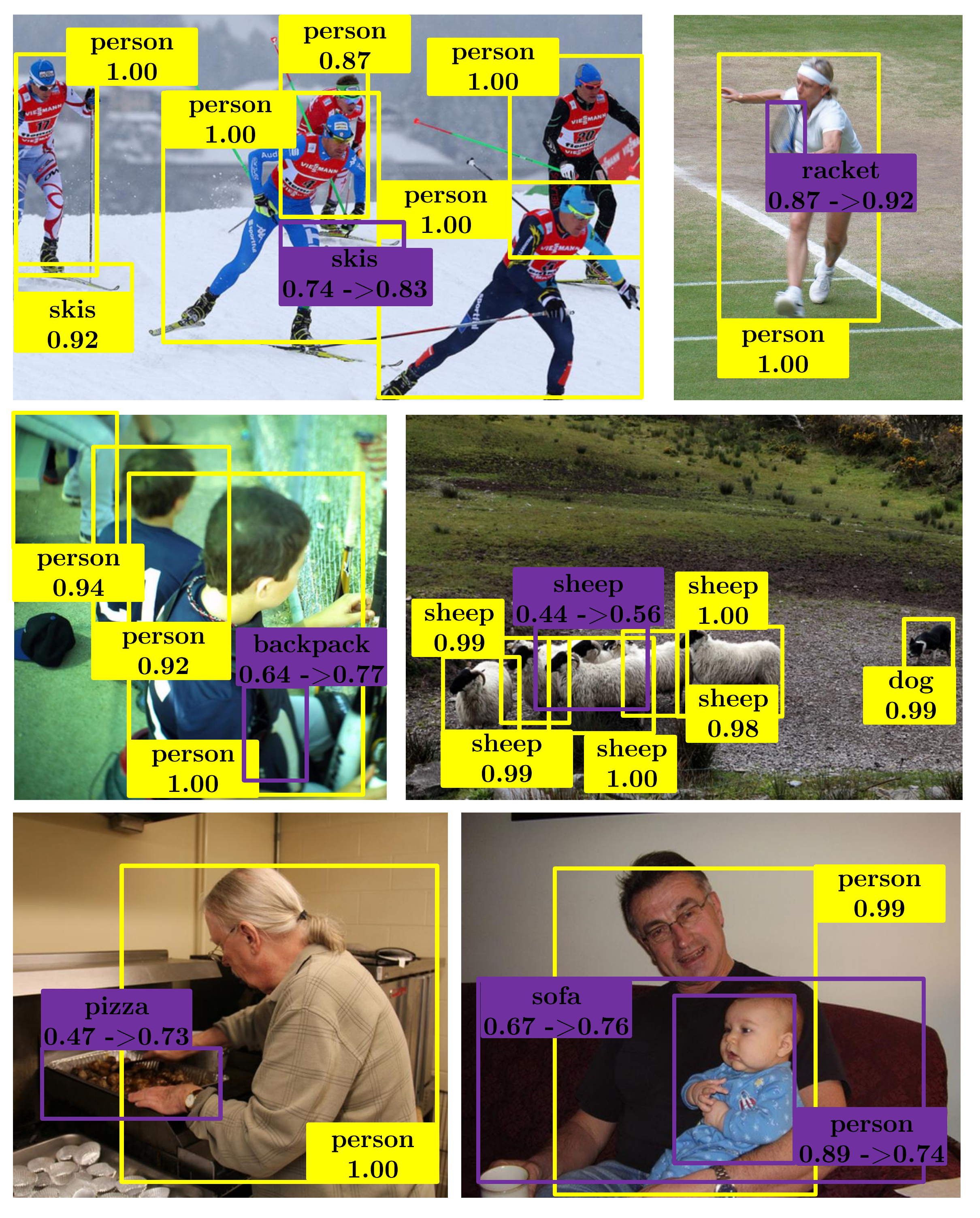}
  \vspace{-0.2in}
  \caption{Four successful reasoning examples and two failure cases. Please see Sec.A1 for a detailed explanation. {\label{fig:prior2}}}
  \vspace{-0.1in}
\end{figure}

We show more qualitative results in Figure~\ref{fig:prior2} to present how the current version of SMN reasons with context for object detection. On the top left, the confidence of {\it skis} is increased due to the detection of {\it person} and their relative location. On the top right, the confidence of {\it tennis racket} is increased despite the motion blur owing to the {\it person} and her pose. Similarly, the {\it backpack} on the middle left gains confidence due to the {\it person} carrying it. On the middle right, we show the example of an occluded {\it sheep} detection. SMN is able to go beyond the overlapping reasoning of NMS, which will prevent the {\it sheep} in the back from being detected. On the bottom row, we show two failure examples, where the potatoes are mistaken as {\it pizza} in the container (left), and the suppression of baby ({\it person}) given the {\it person} holding it.

\begin{table*}[!t]
\centering
\renewcommand{\arraystretch}{1.2}
\renewcommand{\tabcolsep}{1.2mm}
\definecolor{LightGreen}{rgb}{0.75,1,0.75}
\definecolor{LightRed}{rgb}{1,0.75,0.75}
\definecolor{LightBlue}{rgb}{0.75,0.75,1}
\caption[caption]{\label{tab:voc2007}{VOC 2007 test object detection average precision. {\bf S} stands for SoftMax based testing, and {\bf H} for HardMax. {\bf $\not$N} is short for Non-aggressive NMS, where top $5$k RoIs are directly selected without NMS. }}
\vspace{-0.1in}
\resizebox{\linewidth}{!}{
\begin{tabular}{@{} C{0.5cm} !{\color{gray}\vrule} L{2.0cm} !{\color{gray}\vrule} C{0.8cm} !{\color{gray}\vrule} C{0.8cm} !{\color{gray}\vrule} x !{\color{gray}\vrule} x x x x x x x x x >{\columncolor{LightRed}}x x x x x >{\columncolor{LightGreen}}x x >{\columncolor{LightRed}}x x x x @{}}
\Xhline{1pt}
$N$ & \textbf{Method} & \textbf{$\not$N} & \textbf{Max} & mAP & aero      & bike      & bird      & boat      & bottle     & bus        & car        & cat        & chair      & cow        & table      & dog        & horse      & mbike      & persn     & plant      & sheep      & sofa       & train      & tv  \\
\Xhline{1pt}
\parbox[t]{2.5mm}{\multirow{8}{*}{\rotatebox[origin=c]{90}{$N{=}5$}}} & Baseline~\cite{chen17implementation} & \xmark & {\bf S} & 65.8 & 66.8 & 71.3 & 66.1 & 50.0 & 42.0 & 74.5 & 79.6 & 79.3 & 42.4 & 76.3 & 58.0 & 77.3 & 79.4 & 69.1 & 70.1 & 42.5 & 60.0 & 64.8 & 75.0 & 72.2 \\
& SMN Full & \xmark & {\bf S} & 66.4 & 66.3 & 75.3 & 65.4 & 53.3 & 42.2 & 74.1 & 79.5 & 81.9 & 44.5 & 72.7 & 61.9 & 76.5 & 77.0 & 69.7 & 70.1 & 41.8 & 63.9 & 65.5 & 75.1 & 71.6 \\
\Xcline{2-25}{0.5pt}
& Baseline & \xmark & {\bf H} & 65.4 & 67.3 & 71.3 & 60.1 & 50.0 & 41.9 & 74.6 & 79.6 & 79.3 & 42.4 & 76.4 & 57.9 & 77.2 & 79.4 & 69.3 & 70.1 & 42.5 & 60.0 & 64.8 & 75.1 & 68.6 \\
& SMN Full & \xmark & {\bf H} & 64.6 & 60.7 & 71.4 & 65.3 & 50.9 & 42.2 & 74.2 & 79.4 & 78.9 & 42.3 & 67.9 & 59.0 & 75.9 & 72.4 & 69.5 & 70.0 & 41.8 & 59.6 & 65.5 & 75.0 & 68.1 \\
\Xcline{2-25}{0.5pt}
& Baseline & \cmark & {\bf S} & 66.0 & 67.4 & 71.2 & 66.8 & 51.5 & 41.7 & 75.3 & 79.8 & 79.1 & 43.3 & 76.7 & 57.6 & 76.6 & 79.6 & 69.8 & 70.3 & 41.2 & 60.3 & 64.7 & 76.0 & 71.5 \\
& SMN Full & \cmark & {\bf S} & 66.6 & 65.5 & 71.4 & 66.1 & 54.6 & 41.7 & 75.2 & 79.6 & 82.4 & 45.8 & 75.4 & 63.1 & 76.6 & 77.6 & 69.2 & 70.5 & 41.0 & 63.7 & 64.8 & 76.0 & 71.1 \\
\Xcline{2-25}{0.5pt}
& Baseline & \cmark & {\bf H} & 65.8 & 67.4 & 71.2 & 66.8 & 51.2 & 41.7 & 75.3 & 79.8 & 79.0 & 43.1 & 76.7 & 57.6 & 76.8 & 79.6 & 69.8 & 70.3 & 41.3 & 60.3 & 64.8 & 75.8 & 68.2 \\
& SMN Full & \cmark & {\bf H} & 65.4 & 60.8 & 71.3 & 66.1 & 51.8 & 41.9 & 75.0 & 79.6 & 78.4 & 43.3 & 75.3 & 62.8 & 76.4 & 78.1 & 69.2 & 70.5 & 40.9 & 59.2 & 64.4 & 75.8 & 67.9 \\
\Xhline{1pt}
\parbox[t]{2.5mm}{\multirow{8}{*}{\rotatebox[origin=c]{90}{$N{=}10$}}} & Baseline & \xmark & {\bf S} & 70.3 & 67.5 & 78.5 & 67.1 & 53.4 & 54.3 & 78.0 & 84.7 & 84.4 & 48.9 & 82.1 & 66.4 & 77.3 & 80.8 & 75.2 & 77.0 & 46.1 & 70.7 & 64.8 & 75.0 & 73.6 \\
& SMN Full & \xmark & {\bf S} & 69.4 & 66.8 & 79.0 & 69.1 & 52.3 & 53.9 & 73.7 & 82.8 & 83.6 & 46.6 & 78.5 & 64.2 & 76.7 & 80.2 & 75.0 & 77.1 & 44.6 & 67.2 & 67.7 & 75.9 & 72.7 \\
\Xcline{2-25}{0.5pt}
& Baseline & \xmark & {\bf H} & 68.0 & 67.5 & 78.5 & 67.1 & 50.4 & 50.3 & 74.9 & 79.9 & 79.4 & 47.0 & 77.0 & 64.8 & 77.3 & 80.9 & 69.7 & 77.0 & 43.4 & 67.0 & 65.1 & 75.0 & 69.0 \\
& SMN Full & \xmark & {\bf H} & 67.8 & 67.0 & 79.0 & 66.6 & 49.8 & 49.8 & 73.8 & 79.8 & 79.6 & 42.5 & 75.9 & 64.1 & 76.7 & 80.2 & 75.1 & 77.0 & 42.6 & 64.7 & 66.4 & 76.0 & 68.5 \\
\Xcline{2-25}{0.5pt}
& Baseline & \cmark & {\bf S} & 70.4 & 67.5 & 79.0 & 67.6 & 55.2 & 53.4 & 78.9 & 84.5 & 84.0 & 49.6 & 82.0 & 63.4 & 80.3 & 80.6 & 75.7 & 77.3 & 44.8 & 66.7 & 65.8 & 78.5 & 73.2 \\
& SMN Full & \cmark & {\bf S} & 70.0 & 68.3 & 78.1 & 69.5 & 55.0 & 53.6 & 77.7 & 85.1 & 82.5 & 49.2 & 78.0 & 63.8 & 76.5 & 80.0 & 76.0 & 77.5 & 44.3 & 67.6 & 66.6 & 78.7 & 71.8 \\
\Xcline{2-25}{0.5pt}
& Baseline & \cmark & {\bf H} & 68.8 & 67.3 & 79.0 & 67.5 & 52.2 & 49.2 & 75.3 & 80.1 & 79.2 & 47.6 & 81.8 & 63.5 & 76.5 & 80.6 & 75.4 & 77.3 & 42.1 & 66.7 & 64.9 & 76.0 & 73.1 \\
& SMN Full & \cmark & {\bf H} & 68.3 & 66.2 & 78.1 & 66.6 & 51.7 & 49.9 & 75.8 & 85.0 & 78.9 & 47.6 & 76.1 & 64.1 & 76.7 & 79.9 & 76.2 & 77.4 & 42.0 & 65.1 & 65.4 & 76.1 & 67.9 \\
\Xhline{1pt}
\end{tabular}
}
\vspace{-0.1in}
\end{table*}

\begin{table*}[t]
\centering
\renewcommand{\arraystretch}{1.2}
\renewcommand{\tabcolsep}{1.2mm}
\caption[caption]{\label{tab:finalvoc}{VOC 2007 test object detection average precision. We use SoftMax and top $5$k RoIs during testing for all the methods compared (except~\cite{ren2015faster}). }}
\vspace{-0.1in}
\resizebox{0.95\linewidth}{!}{
\begin{tabular}{@{} L{2.5cm} !{\color{gray}\vrule} x !{\color{gray}\vrule} *{20}{x} @{}}
\Xhline{1pt}
\textbf{Method} & mAP & aero      & bike      & bird      & boat      & bottle     & bus        & car        & cat        & chair      & cow        & table      & dog        & horse      & mbike      & persn     & plant      & sheep      & sofa       & train      & tv  \\
\Xhline{1pt}
FRCNN~\cite{ren2015faster} & 70.0 & 68.7 & 79.2 & 67.6 & 54.1 & 52.3 & 75.8 & 79.8 & 84.3 & {\bf 50.1} & 78.3 & 65.1 & {\bf 82.2} & {\bf 84.8} & 72.9 & 76.0 & 44.9 & 70.9 & 63.3 & 76.1 & 72.6 \\
\Xhline{0.5pt}
Baseline~\cite{chen17implementation} & {\bf 71.2} & 67.6 & 78.9 & 67.6 & 55.2 & {\bf 56.9} & {\bf 78.8} & {\bf 85.2} & 83.9 & 49.8 & {\bf 81.9} & 65.5 & 80.1 & 84.4 & 75.7 & 77.6 & {\bf 45.3} & 70.8 & 66.9 & 78.2 & {\bf 72.9} \\
MLP & 70.9 & {\bf 71.7} & 80.0 & {\bf 70.9} & {\bf 60.0} & 56.6 & 78.2 & 85.0 & {\bf 85.5} & 47.5 & 72.7 & 64.2 & 76.6 & 83.5 & {\bf 75.8} & {\bf 77.8} & 45.2 & {\bf 72.3} & {\bf 68.1} & 76.3 & 70.4 \\
SMN & 71.1 & 67.1 & {\bf 81.2} & 70.3 & 55.5 & 54.0 & 78.3 & 85.1 & 83.7 & 49.4 & 80.9 & {\bf 66.1} & 80.1 & 83.5 & 75.7 & 77.7 & 45.1 & 69.7 & 67.1 & {\bf 78.4} & 72.6 \\
\Xhline{1pt}
\end{tabular}
}
\vspace{-0.1in}
\end{table*}

\section*{A2. Category-wise Ablative Analysis on VOC}
Due to space limit, we excluded category-wise numbers in the main paper. However, we believe it is interesting to check the category-wise numbers and get a more insightful idea for the ablative analysis part (Sec.~6.2, main paper).

In Table~\ref{tab:ablative}, we listed our ablative analysis on different training strategies, but the best mAP we can reach on VOC is still behind the baseline: for $N{=}5$ it is $64.6\%$ compared to $65.8\%$; for $N{=}10$ it is $67.8\%$ compared to $70.3\%$. Judging from the COCO AR metrics, we speculate the issue lies in recall. This motivates us to take the best SMN Full model and conduct an investigation specifically targeting recall.

Turns out, the biggest issue lies in the {\bf SoftMax \vs HardMax} strategy. For SMN, we initially deployed a HardMax one: given a bounding box, we proceed with the most confident class (and take the bounding box after regression corresponding to that particular class). This ignores all the rest classes that are potentially competitive. \Eg, on COCO {\it snowboard} is usually confused with {\it skis} when buried in the snow; {\it hot-dog} is usually confused with {\it pizza} when held in a person's hands. Note that this does not cause a problem for NMS, because the de-duplication is done in a per-class manner. Therefore, for a confusing bounding box where \eg the belief for {\it laptop} is $50\%$ and {\it keyboard} is $35\%$, NMS can keep both candidates in the top $N$ detections, whereas for SMN {\it keyboard} is suppressed because {\it laptop} has a higher confidence. While in theory this is not an issue because SMN can {\it learn} to revisit the same region, we find it extremely unlikely to happen in practice, mainly due to the heavy burden on SMN to learn de-duplication automatically and may also attribute to our current feature design. To investigate whether it is indeed the case, we added two ablative experiments: a) using HardMax strategy for baseline with NMS, meaning an initial proposal can only be selected {\it once} -- by the most likely class; and b) using SoftMax for SMN where the final $N$ detections can come from all classes of the $N$ bounding boxes returned by sequential prediction.

As shown in Table~\ref{tab:recall}, we find it worked in both ways: the recall indeed boosts for SMN when a SoftMax strategy is used; and removing the confusing categories for the same bounding box hurts the recall for NMS. For COCO, the improvement on AR metrics directly reflects this finding, however it is less obvious for VOC. Here we additionally include evidence from category-wise results in Table~\ref{tab:voc2007} to corroborate the observation. For example, categories like {\it cow} and {\it sheep} get consistent improvements in SMN since they are more likely to confuse, where as distinctive categories like {\it person} almost remain the same. Overall SoftMax outperforms HardMax in most cases.

Normally, $k{=}300$ RoIs are selected by NMS (\ie region proposal) before feeding into region classification. However, SMN as a sequential prediction method is more vulnerable to such an aggressive region selection scheme, because one miss along the chain can negatively impact all the follow-up detections. Therefore, in addition to the two strategies, we also include the analysis on the impact of the number of regions sampled. Specifically, we include a non-aggressive NMS scheme, where the top 5$k$ proposals are directly selected without NMS. 

Note that because the baseline feeds more RoIs ($k{=}300$ or $k{=}5,000$) for final evaluation, it still bears a subtle advantage over SMN when testing with HardMax. For example, if bounding box $B_1$ suppresses {\it cow} over {\it sheep}, there is still chance that a nearby (\eg meansured by IoU) bounding box $B_2$ where {\it sheep} is selected over {\it cow}. On the other hand, SMN gets $N{\ll}k$ chances for picking the candidates. This difference is reflected when we compare $k{=}300$ \vs $k{=}5,000$ for baseline: SoftMax based testing has a larger margin over HardMax when $k$ is smaller. Regardless of this, SMN is still able to achieve on-par ($N{=}10$) or better ($N{=}5$) results in terms of mAP.

\section*{A3. Final Results on VOC}
We also excluded the final results on VOC comparing 1) the baseline FRCNN, 2) the MLP where a $5$-layer ConvNet is directly stacked on top \texttt{conv5\_3} for context aggregation, and 3) our SMN. We report the results here. For the final evaluation, we use top $5$k RoIs for region sampling and the SoftMax strategy for all methods. Due to the memory limitation, the same idea of first using NMS to find easy examples, storing them in the spatial memory and then predicting with SMN is used (Sec.~\ref{sec:moreiter}). For VOC we set $N_1{=}10$ and bootstrap from a $N_2{=}10$ SMN model, so in total $N{=}20$ bounding boxes are sent for evaluation. 

The result can be found in Table~\ref{tab:finalvoc}. As can be seen, our method is on-par with baseline and MLP (${\sim}71\%$ mAP). This difference to COCO is reasonable since compared to COCO, there is not much ``juice'' left for context reasoning, in terms of both {\it quantity} (number of images to train SMN on top of FRCNN) and {\it quality} (how difficult the detection of objects are in the scene).

\end{document}